\documentclass[a4paper]{article}
\usepackage[utf8]{inputenc}
\usepackage[left=1.5cm, right=1.5cm, top=2cm, bottom=1.5cm]{geometry}
\usepackage{authblk}
\usepackage{url}
\usepackage[colorlinks=true,linkcolor=blue,citecolor=blue,urlcolor=blue]{hyperref}
\usepackage[pdftex]{graphicx,color}	%graphics
\usepackage{multirow}
\usepackage{soul}
\usepackage{amsmath}

\begin{document}

\title{Machine learning augmented diagnostic testing to identify sources of variability in test performance.}
\author[1]{Christopher J.\ Banks}
\author[1]{Aeron Sanchez}
\author[2]{Vicki Stewart}
\author[2]{Kate Bowen}
\author[3]{Thomas Doherty}
\author[4]{Oliver Tearne}
\author[5]{Graham Smith}
\author[1,6,*]{Rowland R.\ Kao}
\affil[1]{\small Roslin Institute, University of Edinburgh, Edinburgh, UK}
\affil[2]{\small UK Farmcare Ltd., Stone, UK}
\affil[3]{\small Department of Mathematics and Statistics, University of Strathclyde, Glasgow, UK}
\affil[4]{\small The Animal and Plant Health Agency, Weybridge, Surrey, UK}
\affil[5]{\small National Wildlife Management Centre, Animal and Plant Health Agency, Sand Hutton, York, UK.}
\affil[6]{\small School of Physics and Astronomy, University of Edinburgh, Edinburgh, UK}
\affil[*]{\small Correspondence: \texttt{rowland.kao@ed.ac.uk}}

%\date{}
\maketitle

%%%%%%%%%%%%%%%%%%%%%%%%%%%%
\begin{abstract}
  Diagnostic tests that can detect pre-clinical or sub-clinical infection, are one of the most powerful tools in our armoury of weapons to control infectious diseases. Considerable effort has been paid to improving diagnostic testing for human, plant and animal diseases, including strategies for targeting the use of diagnostic tests towards individuals who are more likely to be infected. We use machine learning to assess the surrounding risk landscape under which a diagnostic test is applied to augment its interpretation. We develop this to predict the occurrence of bovine tuberculosis incidents in cattle herds, exploiting the availability of exceptionally detailed testing records. We show that, without compromising test specificity, test sensitivity can be improved so that the proportion of infected herds detected improves by over 5 percentage points, or 240 additional infected herds detected in one year beyond those detected by the skin test alone. We also use feature importance testing for assessing the weighting of risk factors. While many factors are associated with increased risk of incidents, of note are several factors that suggest that in some herds there is a higher risk of infection going undetected.

  \paragraph{Author Summary}
  Bovine tuberculosis (bTB) remains a major challenge for cattle farming in Great Britain, causing significant economic and animal welfare impacts. The standard skin test used to detect bTB is highly specific but can miss some infected herds. In this study, we used machine learning to combine detailed national testing records with herd information, creating a model that improves the detection of infected herds. Our approach increases the proportion of infected herds identified by over 5 percentage points---equivalent to 240 additional herds detected in one year---without increasing the number of false positives. Alternatively, if the model is tuned to focus on specificity, it can reduce unnecessary restrictions on over 5,000 herds that are not truly infected. We also used a simulation model to show that these improvements could potentially reduce the number of infected animals and outbreaks in high-risk areas over time. Our results demonstrate that machine learning can enhance existing disease testing strategies, offering practical benefits for disease control and farming communities.
\end{abstract}

%%%%%%%%%%%%%%%%%%%%%%%%%%%
\section*{Introduction}
Diagnostic tests are an essential tool in the armoury of infectious disease control. Improving diagnostic test performance is therefore of considerable research interest, usually centring on improvements in the test itself, or in finding alternative testing approaches, especially when the primary screening test is of sub-optimal sensitivity. This is the case for bovine tuberculosis (bTB) in Great Britain (GB) and Ireland, where, despite considerable effort, eradication of the disease has been elusive. bTB is the most economically important livestock disease in GB, costing around £100m annually and having a substantial impact on animal health and welfare, the trade and supply of livestock, and the livelihoods of farmers~\cite{godfray2013,DepartmentforEnvironmentFoodandRuralAffairs2021}. bTB is caused by the bacterium \emph{Mycobacterium bovis} and results in a chronic, primarily respiratory disease. Prevalence of the disease in GB varies, being especially high in the  high-risk area (HRA) and ``edge'' areas of the South West of England and in parts of Wales, while Scotland is officially bTB-free~\cite{Bessell2012,Uberoi2019}.

While infection of Eurasian badgers is implicated in bTB persistence in large parts of GB and Ireland, the intractable nature of the problem has led to increased interrogation of the performance of the statutory testing tool, the ``skin'' or Single Intradermal Comparative Cervical Tuberculin (SICCT) test. The SICCT test is used for routine surveillance for bTB in the UK, Ireland, Portugal, and increasingly in France. Similar tests are used successfully around the world~\cite{OBrien2023}. While the SICCT test is known to have only moderate sensitivity~\cite{Karolemeas2012}, thus generating false negative results, the test is also highly specific~\cite{Goodchild2015} with a very low proportion of false positive results. High specificity is particularly important when the volume of testing is very high, and so it is a useful tool for detecting and controlling bTB in cattle herds~\cite{DeLaRua-Domenech2006}. Once an infected herd is detected through routine or targeted SICCT testing, or by routine post-mortem meat inspection in abattoirs (slaughter surveillance), and a bTB incident (hereafter referred to as ‘breakdown') has been declared in that herd, the diagnostic sensitivity of the SICCT test is enhanced by short-interval (60-day) follow-up testing, lowering the test-positive cut-off point, occasional removal of non-reactor animals as direct contacts, and supplementation of the SICCT test with parallel interferon gamma (IFN-$\gamma$) blood testing.

Powerful ``machine learning'' data analytical techniques are increasingly being used in epidemiology due to the complex and extensive nature of the data that are often available to analyse epidemiological risks. In the case of bTB, the exceptional density of information on the cattle population, including diagnostic testing results, provides an opportunity to augment skin test interpretation through the inclusion of herd-specific epidemiological information. Stanski et al.~\cite{Stanski2021} established the viability of this approach and showed substantial improvements in test performance at the herd level. Here, we extend this work, introducing several risk factors known to be important for herd risk, including the type of herd and the situation in which a test is taken. Variation in performance across private veterinary practices delivering the bTB testing programme has previously been identified as potentially substantial~\cite{NorthernIrelandAuditOffice2016} and so we also quantify, for the first time, possible variation that is correlated to the veterinary practice conducting the test. While this analysis cannot identify causes of that variability, it could provide indicators of points of further interrogation in the field. 

We also include the local wild badger abundance and, as the supplementary IFN-$\gamma$ blood test constitutes an increasing proportion of statutory testing in herds sustaining bTB breakdowns, we consider this in our evaluation. As IFN-$\gamma$ testing was, until July 2021, also a potential indicator of trial badger culling areas in the HRA of England, this serves as a proxy for variation due to culling related changes in risk~\cite{TBHub2021}.

The objectives of this study are three-fold: (1) to refine the model of Stanski et al.~\cite{Stanski2021}, including further risk factors known to be associated with bTB breakdowns; (2) to assess the potential impact of such a model being used ``in the field'' using a simulation; and (3) to identify the risk factors that the model finds most important in predicting bTB breakdowns.

\section*{Methods}
\subsection*{Data curation}
Data for this analysis were extracted from the Animal and Plant Health Agency (APHA) bTB surveillance database (SAM)~\cite{AnimalandPlantHealthAgency} and the Cattle Tracing System (CTS)~\cite{BritishCattleMovementService} databases. Variables relating to SICCT tests, farm characteristics, and animal movements were compiled into a single dataset that was used to train a machine learning algorithm. Each record of the resultant dataset relates to one SICCT test event as recorded in SAM, along with all metadata (risk features) relevant to that test compiled from SAM and CTS. Data on the veterinary practice conducting the test and the tuberculin batches used were supplied by UK Farmcare~\cite{UKFarmcareLtd.}.

The resultant dataset comprises every recorded SICCT test event between January 2012 and September 2021 for the whole of GB (1.3m records), along with metadata as in Table~\ref{tab:metadata}. Each of these metadata comprises a feature of the model and Table~\ref{tab:metadata} records their datatype. All of the features are time-variant, each piece of metadata being a property of the herd at the time of testing; different tests (at different times) on the same herd may have different values for each feature.

\begin{table}
  \centering
  \begin{tabular}{|l|l|}
    \hline
    Feature & Datatype \\
    \hline  
    Herd-level result of the test & Boolean\\
    Date of the test & Date\\
    Month of the year in which the test occurred & Number\\
    Whether the severe interpretation was applied at testing & Boolean\\
    Number of animals tested & Number\\
    Holding location (Northing) & Number\\
    Holding location (Easting) & Number\\
    Results of previous SICCT tests in the same herd & Boolean\\
    Results of 2nd last previous SICCT tests in the same herd & Boolean\\
    Number of days since the last test in the same herd & Number\\
    Number of days since the herd last entered breakdown & Number\\
    Number of prior IFN-$\gamma$ tests conducted on the herd & Number\\
    Test type (routine, pre-movement, etc.) & Categorical \\
    Type of herd (dairy, beef, etc.) & Categorical\\
    Cattle moved into herd in the last 90 days & Number\\
    Cattle moved into herd in the last year & Number\\
    Cattle moved into herd in the last 2 years & Number\\
    Cattle moved into herd in the last 4 years & Number\\
    Cattle moved out of herd in the last 90 days & Number\\
    Cattle moved out of herd in the last year & Number\\
    Cattle moved out of herd in the last 2 years & Number\\
    Cattle moved out of herd in the last 4 years & Number\\
    Cattle moved into herd in last 90 days from farm with breakdown within 2 years & Number\\
    Cattle moved into herd in last year from farm with breakdown within 2 years & Number\\
    Cattle moved into herd in last 2 years from farm with breakdown within 2 years & Number\\
    Cattle moved into herd in last 4 years from farm with breakdown within 2 years & Number\\
    Cattle moved out of herd in last 90 days from farm with breakdown within 2 years & Number\\
    Cattle moved out of herd in last year from farm with breakdown within 2 years & Number\\
    Cattle moved out of herd in last 2 years from farm with breakdown within 2 years & Number\\
    Cattle moved out of herd in last 4 years from farm with breakdown within 2 years & Number\\
    APHA bTB herd risk score & Number\\
    Mean badger abundance at the holding location & Number\\
    Veterinary practice that conducted the test & Categorical\\
    Tuberculin batch number used for the test (bovine/avian) & Categorical\\
    \hline
  \end{tabular}
  \caption[Test metadata]{Test record metadata used as features in the model, along with their datatype. All features can be considered time-variant, with each being a property of the herd at the time of the test.}
  \label{tab:metadata}
\end{table}

Veterinary practice data includes 117,411 of the tests conducted, covered by 404 practices. Tuberculin batch data includes 57,689 of the tests conducted, with 661 bovine tuberculin batches and 646 avian tuberculin batches recorded.

For each test we also record whether the herd lost its Officially Tuberculosis Free (OTF) status following the detection of test reactors (a bTB herd breakdown) and at least one of those reactors presented with typical lesions of bTB at slaughter and/or positive culture results within 90 days of that test. This is the outcome (gold standard) to be predicted by the model. It is measured with respect to a breakdown being confirmed within 90 days of the test being considered, either because of that test or of a subsequent test within 90 days (e.g. via a pre-movement test, required when cattle are sent to another premises), and where a lesion- or culture-positive reactor is identified prior to the herd regaining its OTF status. 

\subsection*{Model training}
The compiled dataset was split into a training and a testing set; the testing set being the tests conducted from the year 2020 onwards (\~15\% of the full dataset). We used a Histogram-based Gradient Boosted Tree (HGBT)~\cite{Ke2017} as the model of choice with a number of factors driving that decision. HGBTs have a number of desirable properties for this particular dataset, most importantly they natively handle missing and categorical data without the need for pre-processing (such as data imputation or one-hot encoding) as the algorithm pre-bins values before training allowing a bin to represent missing data or a set of bins to represent categories. Previously, Stanski et al.\cite{Stanski2021} tested a number of algorithms on a similar dataset and found Gradient Boosted Trees to be the best performing. For our new dataset on the same range of models (Support Vector Classifiers, Neural Networks, Random Forests, Gradient Boosted Trees) the HGBT was the best performing. We used the HGBT implementation from Scikit-learn in Python~\cite{scikit-learn} (version 1.3.2).

The training of a HGBT model requires a number of ``hyperparameters'' to be tuned, which change the constraints on model structure and complexity. In our model we tune the ``max leaf nodes'' and ``max depth'', which affects the complexity of decision trees computed, and ``learning rate'', which alters the weighting of successive decision trees to balance computation time and reduce overfitting. We used a random search with 200 samples over the parameter space of ``learning rates'' between $0.01$ and $1.0$, ``max leaf'' nodes between $2$ and $150$, and ``max depths'' between $3$ and $25$. Other hyperparameters are as the default in the implementation.

As the features are time-variant in nature we use Temporal Cross-validation to fit hyperparameters. Temporal cross-validation always uses the most recent data as the validation set, ensuring the validation set is always future data with respect to the training set, thus the length of the dataset increases in each successive fold. Here we use five temporal splits, meaning five different lengths of data, always keeping the final portion as the test set. Thus, in combination with the random-sampled hyperparameter fitting, we take 1000 samples to train the model, choosing the best performance over the reserved (post 2020) test set. Performance is measured using the area under the receiver operating characteristic curve (AUROC). The final model has a ``learning rate'' of $0.06$, a ``max leaf nodes'' of 120, and a ``max depth'' of 16.

The trained model takes each of the risk features as input and returns the predicted probability of a confirmed breakdown. To make a binary classification, a decision threshold is taken on the returned probability. The choice of this decision threshold allows the tuning of the balance between sensitivity and specificity.

\subsection*{Simulation model}
To quantify the possible effect that application of the model could have on bTB transmission within the British cattle population, we made use of an existing individual-based model for simulating and predicting bTB spread across two large contiguous areas.

The simulation model allows the following discrete individual animal states: all cattle are born susceptible (S), when infected enter a non-infectious but test sensitive state (T), and then proceeding on at a fixed rate to an infectious stage (I). The sensitivity of the test is fitted separately for each of the two infected states. In addition, infectious stage cattle are assumed to seed the local environment, allowing for continued transmission after cattle are moved away or slaughtered. This environmental reservoir incorporates the role of badgers. Infection between cattle and badgers is assumed to be uniform within geographically defined hexagonal tiles, with each tile having single badger density estimate (i.e. uniform in each tile). Dynamics of infection in badgers are explicitly included in the model fit at the badger social group (or “group”) level. The local density of groups is imputed based on estimates of badger main sett densities at a 500m $\times$ 500m square grid resolution, as previously calculated by Croft et al.~\cite{Croft2017}.

The two areas tested were chosen to represent different bTB risk conditions, with one located in the Derbyshire (Edge Area of England), containing 1431 herds, and the other Devon (High Risk Area), containing 1001 herds. These areas were chosen as they both contain relatively dense cattle populations and bTB levels but are under different epidemiological regimes. The model parameters were fitted in each location using an ABC-SMC method~\cite{Toni2009}, these parameters included the individual level skin-test sensitivity and specificity.

The individual-level skin test sensitivity and specificity in the simulation model are varied to match the \textit{HSe} and \textit{HSp} provided by the selected HGBT model. For each level of individual \textit{Se}/\textit{Sp} we ran the simulation 30 times with the other input parameters being randomly sampled from the fitted posteriors. The \textit{HSe/HSp} values were taken as the mean of all runs. Further details of the simulation model implementation can be found in the supplementary information. With the simulation model fit that matches the HGBT HSe/HSp we assess the number of breakdowns, confirmed breakdowns, and number of individual cattle that test positive (reactors) in the scenario.

As this part of the study makes use of a pre-existing simulation model, the further details of the simulation model are recorded in Supplementary Information.

\subsection*{Risk factor importance}
To identify the risk factors that had a significant impact on the accuracy of the model we used the Shapley Additive Explanations (SHAP) framework~\cite{Lundberg2017}. SHAP assesses the accuracy impact of removing each feature from the model, whilst also taking into account the inherent complex correlations between each of the features in the dataset. We used the mean absolute SHAP value for each feature as a measure of feature importance. The absolute SHAP values for each feature were tested for statistical significance by comparing each feature with a feature composed of a uniform random set of number, using a Mann-Whitney U test (cutoff $p<0.01$).

\section*{Results}
\subsection*{Model training}
The tuned and cross-validated model is able to predict a confirmed breakdown within 90 days of a testing event with 86.1\% accuracy. The Receiver Operating Characteristic (ROC) curve (Figure~\ref{fig:roc}a) shows the balance between herd-level sensitivity/specificity (\textit{HSe/HSp}) for the model in comparison with the \textit{HSe/HSp} of the SICCT test result alone. The area under the ROC curve is 0.90, where 1.0 is a perfect model and 0.5 is no better than a random outcome. The model achieves better overall accuracy at all levels of \textit{HSe/HSp} than the SICCT test result alone.

\subsection*{Model performance}
\begin{figure}
  \centering
  \fbox{(a)\includegraphics[width=0.33\textwidth]{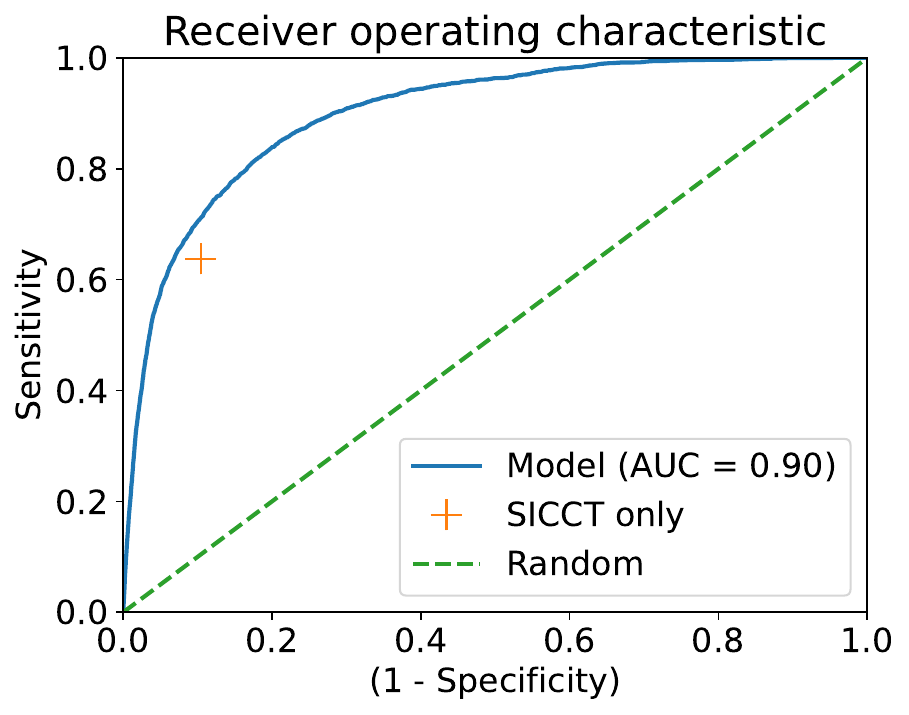}} %\hspace{1em}
  \fbox{(b)\includegraphics[width=0.53\textwidth]{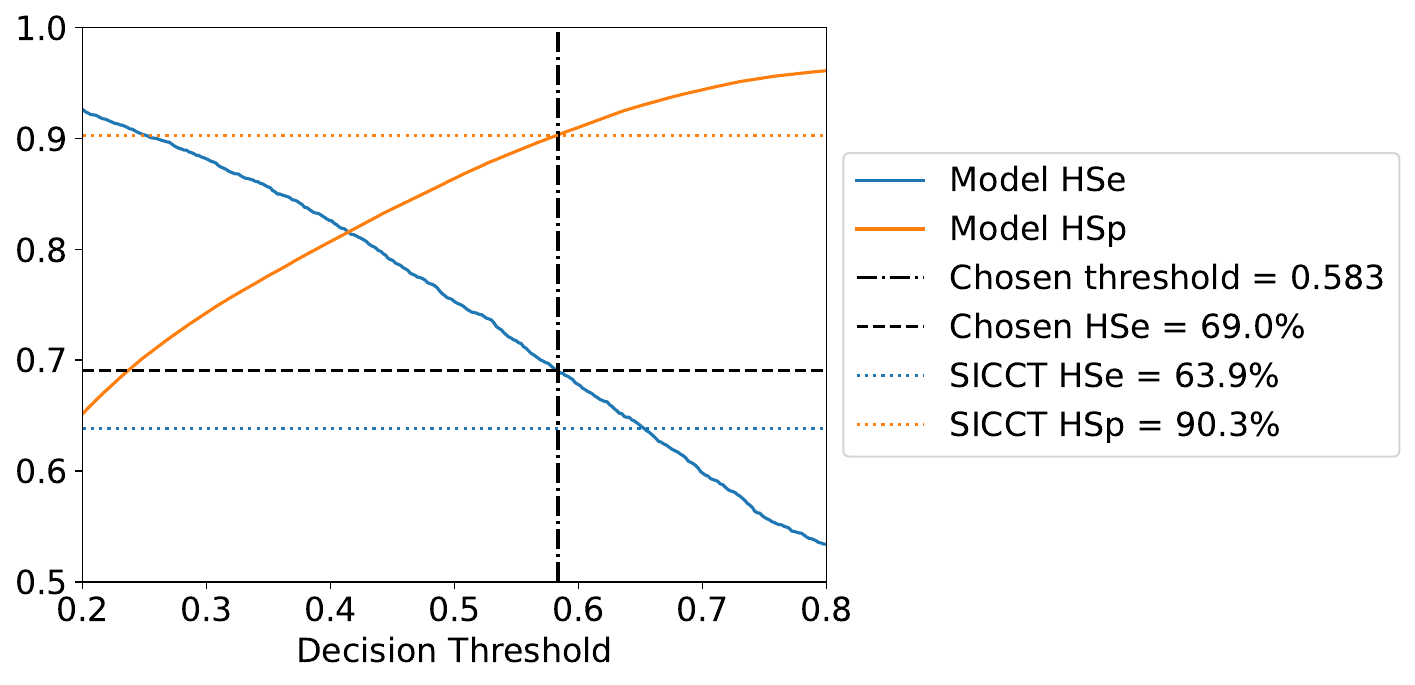}}
  \caption{(a) Receiver operating characteristic (ROC) curve for the diagnostic model. Performance is consistently better than SICCT testing alone for all decision thresholds. 
  (b) The decision threshold choice, such that the herd-level specificity (HSp) is maintained at the level of the SICCT test and the herd-level sensitivity (HSe) is maximised.}
  \label{fig:roc}
\end{figure}

The choice of decision threshold results in different outcomes in the balance between \textit{HSe} and \textit{HSp}. The decision threshold (0.583) was chosen to match the \textit{HSp} of the SICCT test (90.3\%) and to maximise \textit{HSe} (Figure~\ref{fig:roc}b). This achieves an \textit{HSe} of 69.0\%, compared to a SICCT \textit{HSe} of 63.9\%, an \textit{HSe} increase of 5.2 percentage points.

\begin{figure}
  \centering
  \fbox{(a)\includegraphics[width=0.46\textwidth]{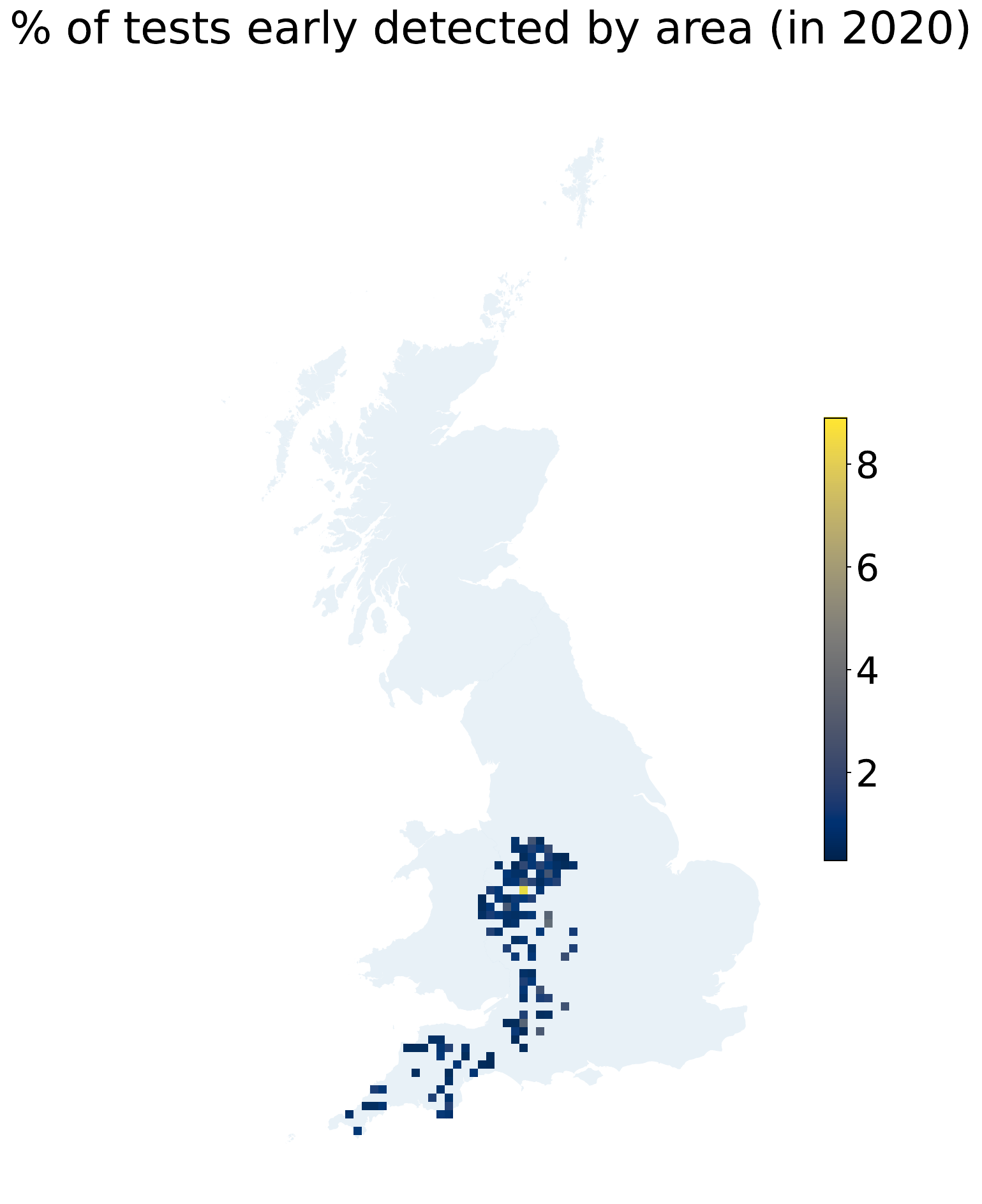}}
  \fbox{(b)\includegraphics[width=0.42\textwidth]{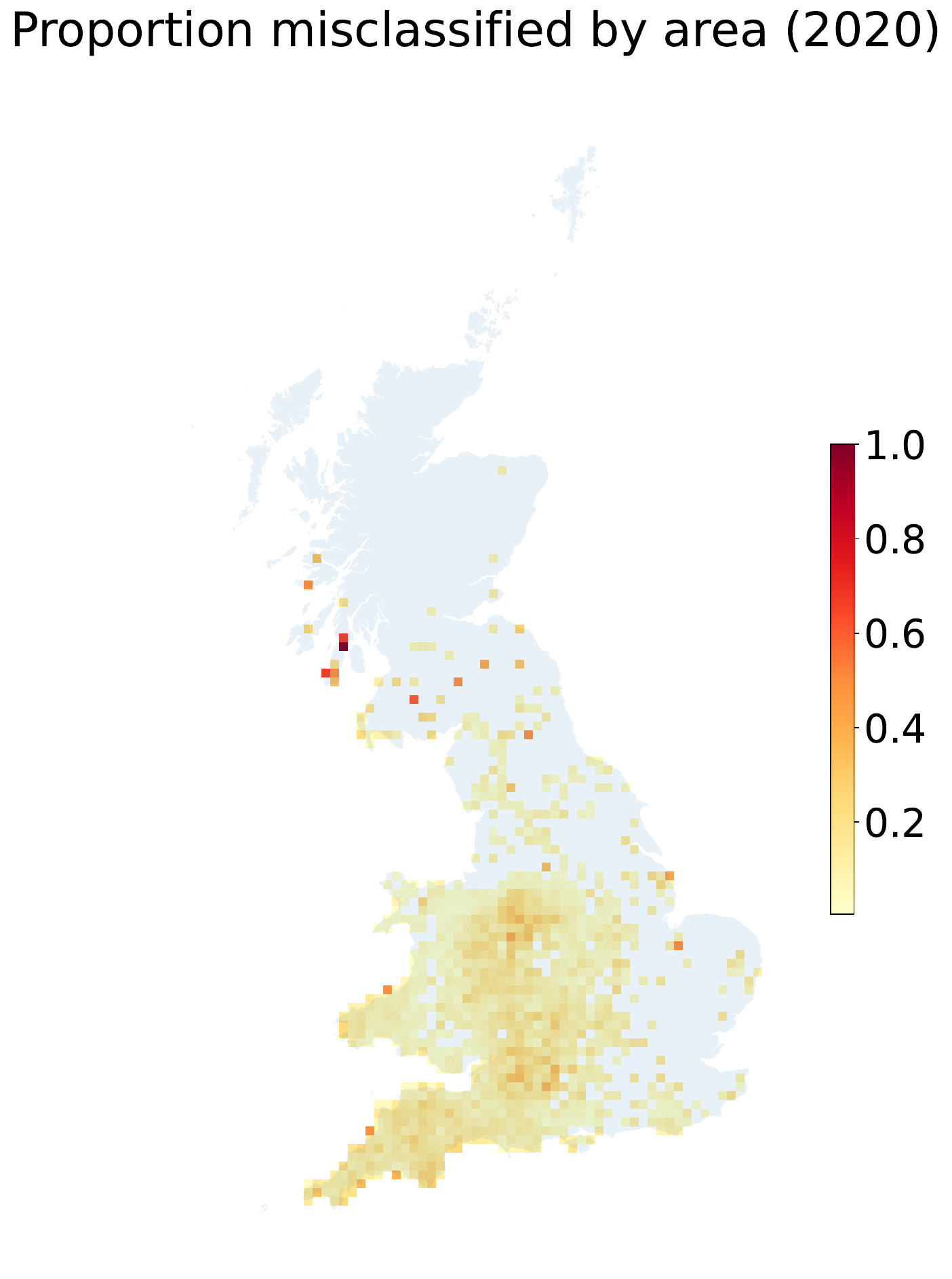}}
  \caption{(a) Proportion (\%) of herds by area that had a negative SICCT test result, but were correctly predicted by the diagnostic model to have a confirmed breakdown, over the year 2020.
  (b) Proportion of herd tests by area that were misclassified by the model in the year 2020.}
  \label{fig:maps_2020}
\end{figure}

Over one year (2020), upon wich the model had not been trained, there were 240 herds that had a confirmed breakdown within 90 days of a SICCT test with negative results. Therefore the model enabled the correct and early identification of those eventual breakdown herds at the time when they initially passed a SICCT test. Figure~\ref{fig:maps_2020}a shows the geographic distribution of those herds that were identified early by the model. These early detected holdings are concentrated in the HRA of England and Wales, but extending into the Edge Area of England, especially around Staffordshire, Derbyshire, and Leicestershire. Figure~\ref{fig:maps_2020}b shows the geographic distribution of holdings misclassified by the model (false positives or false negatives) in the year 2020; these are distributed more uniformly across the country, peaking in areas with low numbers of tests conducted overall.

During the year 2020 the model achieves an \textit{HSe} of 69.0\% and an \textit{HSp} of 90.3\%, compared to a SICCT \textit{HSe} of 63.9\% and SICCT \textit{HSp} of 90.3\%. Table~\ref{tab:conf_mats} compares the confusion matrices for the model and for SICCT testing alone for the year 2020. In this year, the model reduces the proportion of falsely identified negative herds by 14.3\%.

\begin{table}
  \centering
  (a)
  \begin{tabular}{|r|rl|rl|}
    \hline
    \textbf{Model} &Positive& &Negative& \\
    \hline
    True  &2,126 &(1.7\%)     &108,894 &(88.1\%) \\
    False &11,660&(9.4\%)    &953 &(0.8\%) \\
    \hline
  \end{tabular}
  \\\vspace{1em}
  (b)
  \begin{tabular}{|r|rl|rl|}
    \hline
    \textbf{SICCT} &Positive& &Negative& \\
    \hline
    True  &1,967 &(1.6\%)     &108,818 &(88.0\%) \\
    False &11,736 &(9.5\%)    &1,112 &(0.9\%) \\
    \hline
  \end{tabular}
  \caption{Confusion matrices for the diagnostic model (a), compared to the SICCT test alone at herd level (b), in the year 2020. Ground truth here refers to whether the herd had a subsequent breakdown confirmed within 90 days of testing (as opposed to true test outcomes or individual disease status).}
  \label{tab:conf_mats}
\end{table}

If instead we choose to fix \textit{HSe} at the level of the SICCT test and use the model to maximise \textit{HSp}, choosing a decision threshold of 0.651, we achieve a \textit{HSp} of 93.0\% compared to a SICCT \textit{HSp} of 89.6\%, an increase of \textit{HSp} by 2.7 percentage points, or 5225 herds declared negative by the model that showed no further evidence of infection beyond the initial test result, in the year 2020.

\subsection*{Simulation results}
The simulation model output illustrates the potential effects of the increase in \textit{HSe} given by the SICCT test augmented by the model, compared with the performance of SICCT testing alone. These results take the mean number of confirmed breakdowns in the 2020 year of the simulation. Increasing the \textit{HSe} from 63.9\% to 69.1\%, maintaining \textit{HSp} the same as the current SICCT test results is equivalent to a modelled increase in individual test sensitivity of 12.2\%. Using this increase in test sensitivity the resulting number of breakdowns and reactors in each area are detailed in Table~\ref{tab:sim-results}.

\begin{table}
  \centering
  \begin{tabular}{|l|r|r|r|}
    \hline
    &Breakdowns&Confirmed breakdowns&Positive reactors\\
    \hline
    Derbyshire -- SICCT only      & 232 [141--334]& 112 [44--198]& 1043 [608--1575]\\
    Derbyshire -- With sensitivity increase& 236 [146--332]& 106 [38--181]& 960[583--1390]\\
 Derbyshire -- With specificity increase& 226[134--325]& 113[43--200]&911[489--1401]\\
    Devon -- SICCT only           & 157 [88--238]& 86 [32-150]& 690 [400--1025]\\
    Devon -- With sensitivity increase& 149 [85--219]& 74 [26--129]& 594 [378--863]\\
    Devon -- With specificity increase& 152[81--227]& 85[30--147]&589 [310--906]\\
 \hline
  \end{tabular}
  \caption{Results of simulating the transmission of bTB within and between herds in two areas (Derbyshire and Devon) using an individual-based simulation model, for the existing testing regime (SICCT only) and as augmented by the diagnostic model (with model). Confidence intervals shown in square brackets.}
  \label{tab:sim-results}
\end{table}

Changes in breakdown numbers are a result of a combination of (i) increasing the number of detected herds where these would be missed by the SICCT test only but infection is cleared without additional measures and (ii) decreasing the number of onward infections that result from early detection of herds that would have had more substantial outbreaks. Here, assuming the increase in \textit{HSe} provided by augmenting the SICCT test with the model, in simulation the number of breakdowns on farms in Derbyshire and Devon see differing results. In Derbyshire, there is a slight increase of 1.7\%, but a decrease of 5.1\% in Devon, over the year 2020. Confirmed breakdowns remain roughly the same in Derbyshire, rising by 0.9\%, but are reduced by 14.0\% in Devon. However, Individual reactors are reduced in both areas with Derbyshire seeing a reduction of 8.0\%, and Devon by 13.9\%.

If instead we choose to keep \textit{HSe} the same and increase \textit{HSp} from 89.6\% to 93.0\%, according to the specificity-focused decision threshold, we observe that there is a decrease in the number of breakdowns and reactors in both locations. In Derbyshire the number of breakdowns decreased by 2.6\%, and by 3.2\% in Devon, while the individual reactors decrease by 12.7\% in Derbyshire, and by 14.6\% in Devon.

\subsection*{Risk factor importance}
\begin{figure}
  \centering
  \fbox{\includegraphics[width=0.97\textwidth]{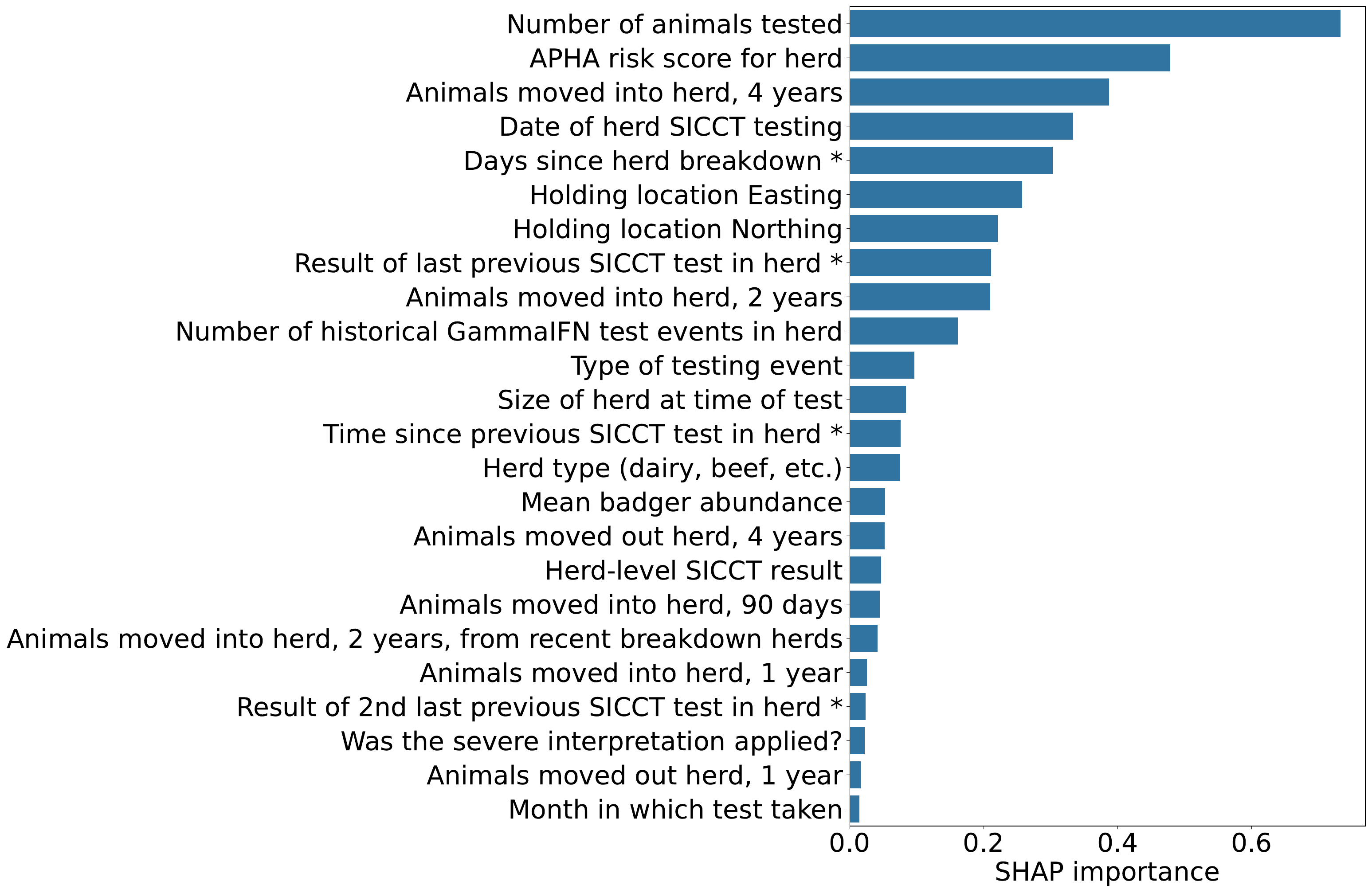}}
  \caption{The relative importance of model features (risk factors), as tested by SHAP importance testing, with a random control variable. Only feature whose absolute SHAP values are significantly greater than the random feature (Mann-Whitney U test, $p<0.01$) are shown. Features marked * refer to previous tests or breakdowns and will be left-censored where the previous test or breakdown is before the first date in the dataset.}
  \label{fig:importance}
\end{figure}

Using SHAP importance testing we do not observe any unexpected results, those factors known to he highly associated with bTB risk appear to make significant contributions in our model. Figure~\ref{fig:importance} shows the importance ranking of each model feature for which the feature's absolute SHAP values are significantly greater than those of a random feature. The most significant features were the number of animals tested, the APHA risk scoring, volume of movements into the herd in the last four years, date of testing, recency of a breakdown in the herd, herd location, and the result of the previous test.

Of those features that can be considered direct risk factors the most important were (in decreasing order of importance): number of animals, movements into the herd, recency of last breakdown, location, and previous test result. Herd size and animal movements were important, with all types of movement contributing more than the random control variable, but movements into the herd over the two to four years prior to the test was the most significant by some margin; other periods and movements out of herd were less significant. Movements from farms with breakdowns within two years appear to make only a small contribution. Neither the veterinary practice conducting the test nor the tuberculin batch had significant importance.

\section*{Discussion}
In this study we have developed a model that predicts the risk landscape associated with bTB breakdowns. We apply this to provide an augmented interpretation of a bTB SICCT test in routine whole-herd testing events. Using a scheme where herd-level specificity is maintained, we are able to improve herd-level sensitivity such that a significant number of false negative test results are caught early in the testing over historical data (Objective 1). By tuning the decision threshold it is also possible to reduce false positives, with that tuning allowing for a balance between the two outcomes. Undetected infections are important because they both potentially extending periods of herd restriction once infection is discovered, and they allow for onward transmission to other herds and to wildlife that would not occur, if detected earlier. On the other hand, because the vast majority of herds do not harbour infection, even small changes in specificity can have a much greater influence on more herds. While the \emph{HSe}-focussed model identifies 240 herds that eventually had confirmed breakdowns but had negative tests, the \emph{HSp}-focussed model identifies 5225 herds that had a positive test but did not eventually have a confirmed breakdown (in 2020). As all breakdowns have a substantial impact simply due to the restrictions that are imposed as a result, this trade-off must carefully consider the approach that would have the greatest positive impact, if implemented in practice. 

% \paragraph{Simulation:}
We further considered this trade-off with a simulation model that looked at the short term onward impact that missed breakdowns have at a regional level (i.e.\ considering on other herds). We further compare two different geographical areas---one in Devon where there has been a decades long bTB problem and the other in Derbyshire, where the history of bTB is more recent. While increasing the \textit{HSe} notably decreases the number of individual reactors in herds in both areas, and thus there is overall a net reduction in infections, the impact on the overall number of breakdowns shows inconclusive benefits, at least in the short term, suggesting that onward benefits impacting the national epidemic may take longer to become manifest. 
Nevertheless, the overall reduction in test reactors in all cases suggests that if applied ``in the field'' in representative areas, this augmented testing method may be successfully used either to reduce impact and burden on farmers and herds. (Objective 2.)

% \paragraph{Risk factors:}
Despite the complexity of the risk factors affecting bTB transmission, the key protagonists are well known and are confirmed in our risk factor analysis. However, one unexpected outcome was the role that date of testing played in model accuracy. This implies that the weighting of other risk factors is not constant, with some becoming more or less important over time in a changing risk landscape. The month in which a test is taken has far lower importance in the model, implying annual seasonality is less important.

We included the veterinary practice and tuberculin batch data because anecdotally these are considered to have some impact on testing outcomes and we wished to test this. Ultimately neither of these features had significant importance in the model. However, the coverage for both batch and veterinary practice were both low; so with improved coverage, either or both could be found to have greater importance (Objective 3.)

% \paragraph{Research in context:}
The usual interpretation (threshold on measurement) of the standard SICCT test is tuned so as to maintain an acceptably low level of false positive results, i.e. to minimise the number of times where herd level restrictions are put into place unnecessarily, where there is no underlying true infection in the herd. This is sensible both because the relatively low overall level of infection across GB (as of Dec 2023, in England a herd prevalence of 4.6\%~\cite{BovineTBstatistics2023}) means that it is mostly truly negative cattle that are being tested, and because the implications of a positive test are severe for both the keeper and the government (i.e. herd restrictions and multiple follow-up tests). While the SICCT test is interpreted in the same fashion across all OTF herds, the epidemiological context is already taken into account in that the frequency of routine bTB herd surveillance testing varies with the regional risk of bTB, and both for where there is already a strong suspicion of infection---bTB breakdown herds that have had their OTF status removed (either suspended/OTFS or withdrawn/OTFW) or for cattled traced from breakdown herds---a more severe interpretation of the SICCT test is automatically used. In this case a larger likelihood of false positive test results is deemed acceptable in exchange for an enhanced probability of removing all infected animals from the herd. Our analysis extends this concept by evaluating whether an assessment of epidemiological risk (i.e. a prior probability that a herd undergoing testing is more or less likely to already harbour infected cattle) could be used to alter the interpretation of the SICCT test, even when the herd in question does not already have its OTF status suspended or withdrawn at the point of testing. That is, even when the herd remains OTF after a clear SICCT performed at the standard interpretation (e.g.\ where only inconclusive reactors are detected), or in those breakdown herds with OTF status suspended in the Low Risk Area of England and in Scotland where the severe interpretation of the SICCT test is not automatically applied. This concept has already been exploited to a more limited degree in Scotland~\cite{Bessell2013} and re-evaluated in England to show proof of concept~\cite{Salvador2018}.

However, those previous results have not considered the full implications that result from either catching some infected herds early or reducing restrictions on others. A machine learning based approach already used to evaluate GB testing~\cite{Stanski2021} provides a platform on which to do a more sophisticated analysis. To accomplish this, we evaluated test results under a specific set of circumstances where subsequent testing is likely to be a reliable indicator of the true infection status of a herd. Specifically, we can evaluate where a historical SICCT test should be viewed differently by looking ahead at whether a confirmed breakdown occurred or not. If a clear or inconclusive test is followed soon after by a positive test with a confirmed reactor, this is a reasonable indicator that there may have been some benefit to the original clear test having been interpreted more severely. Similarly, if an inconclusive test is followed up soon after by a clear set of follow-up tests, this is an indicator that the herd could have safely been relieved of further restrictions beyond the initial culling of test-positive animals, thereby substantially reducing the burden to both the farmer, and the veterinary testing services. 

%\paragraph{Practical implications:}
Whether machine-learning augmentation of testing could be used in practice depends on many factors outside this model including the regulatory framework; we show that in principle such an approach has potential merit and warrants further investigation.  By testing this proposition across the entire record of herd tests using well-established machine learning techniques, we show that simple indicators of epidemiological risks can result in both substantial numbers of herds where an incipient infection can be identified early, as well as many herds that could have been saved a period of restriction and further testing. This approach could potentially be broadened to any disease diagnostic situation where sufficient data is available and could be particularly useful for non-statutory diseases where there is a substantial need for improved testing and greater flexibility in farmer-led management of the disease, as with bovine viral diarrhoea and Johne's disease.

%\paragraph{Extension to measurement threshold:}
Should the necessary data become available, further work could explore in detail a direct interpretation of the skin test measurement itself, rather than the binary outcome in the current model. This could provide an alteration of the detection threshold used for the SICCT test which, in combination with the available epidemiological information, would be used to help decide on the withdrawal of OTF status. This would apply in much the same way as the current severe interpretation, but making use of many more varied risk factors. Any proposed change in testing regime or policy would, of course, need to be subject to further analysis.

\section*{Conclusion} We have shown with historical data that a substantial benefit can be gained by considering complex epidemiological interactions in a single model. Further exploration of sensitivity/specificity benefit trade-offs could also result in improved outcomes. While the underlying analytical model is sophisticated, its outcomes could be easily implemented in real time with software held on a hand held device, and the risk factors displayed graphically in such a way that the results could be useful in the development of a more targeted testing policy, and supplied directly to veterinarians and farmers. 

\section*{Dedication}
\paragraph{\textit{In memoriam} Thomas Patrick Doherty 1975--2024}
The simulation model section of this paper could not have been achieved without the groundwork laid by Tom, the key developer of the simulation model used in this paper.

Tom was a lynch-pin member of our team and the most hard-working, dedicated, kind, and helpful gentleman one could ever hope to meet. Tom sadly passed away before the conclusion of this paper, but his work will live on and continue to help us. He will be sorely missed by all of us. Rest in peace, Tom.

\section*{Funding sources}
This work was supported by the GB bovine TB research budget (grant SE3330, awarded to RRK) held and administered centrally by Defra on behalf of England, Scotland and Wales, and also by the Roslin Institute ISP2(theme 3), BBSRC (grant BBS/E/D/20002174, awarded to Roslin Institute, funds RRK and CJB). Defra approved release of the manuscript, but otherwise the funders had no role in study design, data collection and analysis, decision to publish, or preparation of the manuscript.

\section*{Data Availability Statement}
Data for this analysis were extracted from the Animal and Plant Health Agency (APHA) bTB surveillance database (SAM)~\cite{AnimalandPlantHealthAgency} and the Cattle Tracing System (CTS)~\cite{BritishCattleMovementService} databases. These data were made available under a Data Sharing Agreement between APHA and the University of Edinburgh. Data on the veterinary practice conducting the test and the tuberculin batches used were supplied by UK Farmcare~\cite{UKFarmcareLtd.}. Data requests may be addressed to \texttt{enquiries@apha.gov.uk} for APHA, and \texttt{admin@ukfarmcare.com} for UKFarmcare.

%%%%%%%%%%%%%%%%%%%%%%% 
%%%%% License    %%%%%%
%%%%%%%%%%%%%%%%%%%%%%%
\section*{License}
For the purpose of open access, the author has applied a Creative Commons Attribution (CC~BY) licence to any Author Accepted Manuscript version arising from this submission.

%%%%%%%%%%%%%%%%%%%%%%% 
%%%%% References %%%%%%
%%%%%%%%%%%%%%%%%%%%%%% 

\bibliographystyle{custom}
\bibliography{diagnostics}

\section*{List of Legends}
\begin{itemize}
\item Supplementary Information: Simulation model methods. (btb-diagnostics-supp-info.pdf)
  \begin{itemize}
  \item Supplementary Table 1: Fitted posteriors of TBMI-lite model parameters, for both Derbyshire and Devon areas.
  \end{itemize}
\end{itemize}

\end{document}